\def\year{2021}\relax
\documentclass[letterpaper]{article} 
\usepackage{aaai21}  
\usepackage{times}  
\usepackage{helvet} 
\usepackage{courier}  
\usepackage[hyphens]{url}  
\usepackage{graphicx} 
\usepackage{natbib}  
\usepackage{caption} 
\usepackage{subcaption}
\usepackage{amsmath}
\usepackage{amssymb}
\usepackage{overpic}
\usepackage{verbatim}

\title{From Black-box to White-box: Examining Confidence Calibration under different Conditions}
\author{
    Franziska Schwaiger \textsuperscript{\rm 1},
    Maximilian Henne \textsuperscript{\rm 1},
    Fabian K{\"u}ppers \textsuperscript{\rm 2},
    Felippe Schmoeller Roza \textsuperscript{\rm 1},
    Karsten Roscher \textsuperscript{\rm 1},
    Anselm Haselhoff \textsuperscript{\rm 2}
    \\
}
\affiliations{

    \textsuperscript{\rm 1}Fraunhofer Institute for Cognitive Systems IKS, Munich, Germany\\
    \textsuperscript{\rm 2}Ruhr West University of Applied Sciences, Bottrop, Germany\\
 	\{franziska.schwaiger, maximilian.henne, felippe.schmoeller.da.roza\}@iks.fraunhofer.de\\
 	\{fabian.kueppers, anselm.haselhoff\}@hs-ruhrwest.de
 	
}

\begin{document}

\maketitle

\begin{abstract}

Confidence calibration is a major concern when applying artificial neural networks in safety-critical applications. Since most research in this area has focused on classification in the past, confidence calibration in the scope of object detection has gained more attention only recently. Based on previous work, we study the miscalibration of object detection models with respect to image location and box scale. Our main contribution is to additionally consider the impact of box selection methods like non-maximum suppression to calibration. We investigate the default intrinsic calibration of object detection models and how it is affected by these post-processing techniques.  For this purpose, we distinguish between black-box calibration with non-maximum suppression and white-box calibration with raw network outputs. Our experiments reveal that post-processing highly affects confidence calibration. We show that non-maximum suppression has the potential to degrade initially well-calibrated predictions, leading to overconfident and thus miscalibrated models.
\end{abstract}

\section{Introduction}
\noindent
Modern deep neural networks achieve remarkable results on various tasks but it is a well-known issue that these networks fail to provide reliable estimates about the correctness of predictions in many cases \cite{Niculescu2005, Guo2018}. A network outputs a score attached to each prediction that can be interpreted as the probability of correctness. Such a model is \textit{well-calibrated} if the observed accuracy matches the estimated confidence scores. However, recent work has shown that these confidence scores neither represent the actual observed accuracy in classification \cite{Niculescu2005, Naeini2015, Guo2018} nor the observed precision in object detection \cite{Kueppers2020}. Calibrated confidence estimates integrated in safety-critical applications like autonomous driving can provide valuable additional information with respect to situational awareness and can reduce the risk of hazards resulting from functional insufficiencies by decreasing the space of unknown unsafe scenarios which is a critical part for the safety of the intended functionality (SOTIF ISO/PAS 21448).



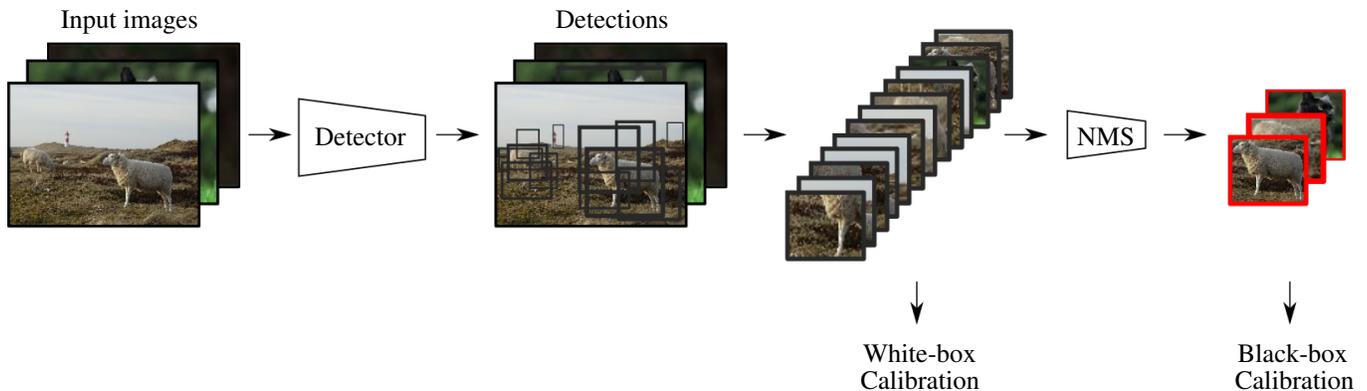
\begin{figure*}
	\centering
	\begin{overpic}[tics=4, width=1.0\linewidth]{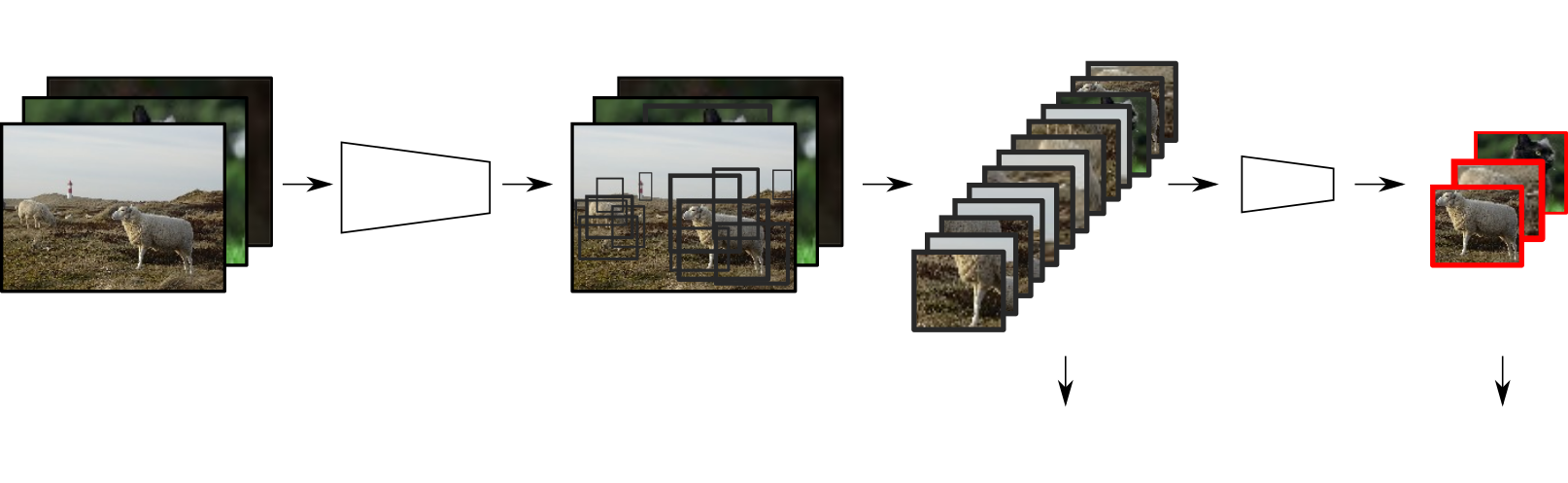}
		\put(23, 18.25){Detector}
		\put(80, 18.25){NMS}
		\put(64, 2){White-box}
		\put(63.8, 0){Calibration}
		\put(92, 2){Black-box}
		\put(91.8, 0){Calibration}
		
		\put(4, 27){Input images}
		\put(41, 27){Detections}
	\end{overpic}
	\caption{Typically a non-maximum suppression (NMS) is applied to all detections of a detection model to fuse and reduce redundant bounding boxes. In our work, we investigate how NMS affects confidence calibration. Thus, we study the difference in calibration before NMS (white-box) and afterward (black-box).}
	\label{img:concept}
\end{figure*}

In the past, most research in this area has focused on classification \cite{Naeini2015, Kull2017, Guo2018, Seo2019, Mukhoti2020}, whereas calibration in object detection has recently gained more attention \cite{Neumann2018, Feng2019, Kueppers2020}. Object detection is a joint task of classification and regression of the predictions' position and scale. Recent work has shown that the regression branch of object detection models also affects confidence calibration \cite{Kueppers2020}. However, the observable detections of a model are commonly processed by non-maximum suppression (NMS) and/or thresholded by a certain confidence score. 
In this work, our goal is to investigate the influence of such post-processing techniques on the model calibration. For this purpose, we adapt common object detection models and examine their miscalibration before NMS on the one hand (white-box scenario). In this way, we have access to the raw predictions of a network and are thus able to examine the network's calibration properties by default. On the other hand, we further apply NMS with increasing intersection over union (IoU) thresholds (black-box scenario), which varies the number of boxes that are suppressed. Changing the parameters of the NMS enables us to examine to what extent the models are intrinsically calibrated and how this is affected by postprocessing techniques. An illustrative representation for the problem setting is demonstrated in Fig. \ref{img:concept}. Furthermore, we use a Faster R-CNN architecture \cite{Ren2015} that uses the cross entropy loss during training and compare it to a RetinaNet \cite{Lin2017} that uses a focal loss. It is already known that models trained with focal loss produce much less confident predictions \cite{Mukhoti2020}. This enables us to further investigate the effect of post-processing methods by comparing the default calibration properties of both model architectures with and without NMS.

This work is structured as follows: we give a review of the current state-of-the-art research in confidence calibration in Section \ref{section:related_work}. We further give a definition of white-box and black-box calibration and a description of our calibration targets in Section \ref{section:main}. In Section \ref{section:experiments}, our experimental results are demonstrated and in Section \ref{section:conclusion} we discuss our findings.

\section{Related Work}
\label{section:related_work}

Numerous methods have been developed in the past to address the miscalibration of neural networks. One of the first representatives of post-processing calibration methods has been histogram binning \cite{Zadrozny2001}, isotonic regression \cite{Zadrozny2002}, Bayesian binning \cite{Naeini2015}, and Platt scaling \cite{Platt1999}, whereas more recently temperature scaling \cite{Guo2018}, beta calibration \cite{Kull2017}, and Dirichlet calibration \cite{Kull2019} have been developed to tackle miscalibration in the scope of classification. In object detection models, dealing with miscalibration presents a different set of challenges and was first addressed by \cite{Neumann2018}, who proposed an additional model output to be utilized as a regularizing temperature applied to the remaining logits. Recently, \cite{Kueppers2020} have studied the effect of position and scale of detected objects to miscalibration and concluded that calibration also depends on the regression output of a detection model. They further provide a framework to include position and scale information into a calibration mapping.

For the task of classification, a common way to measure miscalibration is to adapt the \textit{expected calibration error} (ECE), proposed by \cite{Naeini2015}, which uses a binning scheme to measure the gap between observed frequency and average confidence. \cite{Kumar2019} show in their work that the common ECE underestimates the true calibration error in some cases and provide a differentiable upper bound called \textit{maximum mean calibration error} (MMCE) that can also be used during model training as a second regularization term. For measuring miscalibration in object detection tasks, an extension of the ECE called \textit{detection expected calibration error} (D-ECE) was proposed by \cite{Kueppers2020}, consisting of a multidimensional binning scheme to assess the miscalibration over all predicted features of an object detection model.

Since the standard cross-entropy loss is prone to favor overly confident predictions, further research directions investigate how to directly obtain well-calibrated models after training. Besides the previously mentioned MMCE, the authors in \cite{Pereyra2017} introduce a regularization term penalizing high confident predictions. In contrast, \cite{Muller2019} show that label smoothing yields good probabilities after training. Recently, \cite{Mukhoti2020} investigate the effects of focal loss, originally proposed as a loss term for RetinaNet \cite{Lin2017}, on confidence calibration. They show that using focal loss in conjunction with an adaptive parameter significantly improves the confidence calibration of classification models. We also observe that using focal loss prevents overconfident predictions in our experiments on the RetinaNet with standard hyperparameters \cite{Lin2017}. While well-known object detection models like Faster-RCNN \cite{Ren2015} commonly tend to output overconfident predictions, the probability scores of a RetinaNet rather underestimate the observed frequency.

\section{Defining Confidence Calibration for Object Detection Models}
\label{section:main}
In this section, we describe the definition of black-box and white-box calibration. The idea behind this distinction is to analyze the impact of bounding-box postprocessing on calibration.

An object detector takes an image as input $x$ and outputs predictions in form of a class label $y \in \mathcal{Y}$ with corresponding confidence score $p \in [0, 1]$ and bounding box $r=(c_x, c_x, h, w) \in \mathcal{R}^J$, with $(c_x, c_y)$ being the center position, $(h, w)$ the box height and width and $J$ the size of the used box encoding.
The authors in \cite{Kueppers2020} propose a confidence calibration that not only considers the confidence score $p$ but also includes the box information $r$.  The performance of a detector is thus evaluated by matching its predictions $(\hat{y}, \hat{p}, \hat{r})$ with the ground-truth annotations, where $m=1$ denotes a matched box and $m=0$ a mismatch. More formally, perfect calibration in the scope of object detection is defined by
\begin{align}
\label{eq:calibration:precision}
& \underbrace{\mathbb{P}(M=1 | \hat{P} = p, \hat{Y} = y, \hat{R} = r)}_{\text{precision given } p, y, r} = \underbrace{p}_{\text{confidence}}, \\ \nonumber
& \forall p \in [0,1], y \in \mathcal{Y}, r \in \mathcal{R}^J.
\end{align}

As the detections rarely match the ground-truth perfectly, true positives (TP, $m=1$) and false positives (FP, $m=0$) are obtained by comparing the IoU to a fixed threshold $\tau$. TP and FP correspond to boxes with $\text{IoU} \geq \tau$ and $\text{IoU} < \tau$, respectively.
The process of inference is commonly followed by a non-maximum suppression since an object detection model outputs a huge amount of mostly less confident and redundant detections. On the one hand, we can consider the definition of calibration given by Equation \ref{eq:calibration:precision} to the raw outputs of a detector without any post-processing. We denote this case as the white-box calibration case for the following of this paper. On the other hand, we can also view the NMS as part of the detector and treat the output of the NMS as our desired calibration target. This is denoted as black-box calibration.

In \cite{Kueppers2020}, the detection expected calibration error (D-ECE) is defined as an extension of the commonly used ECE \cite{Naeini2015} for object detection tasks. The D-ECE also includes the box information $r$ by partitioning the space of each variable $k$ into $N_k$ equally spaced bins. The total amount of bins is given by $N_{total} = \prod^K_{k=1} N_k$ and the D-ECE is defined as
\begin{equation}
\text{D-ECE}_K = \sum^{N_{total}}_{n=1} \frac{|I(n)|}{|D|} \cdot |\text{prec}(n) - \text{conf}(n)|,
\end{equation}
where $I(n)$ is the set of all samples in a single bin and $|D|$ the total amount of samples, while $\text{prec}(n)$ and $\text{conf}(n)$ denote the average precision and confidence within each bin, respectively. We use this metric to measure miscalibration in both cases: For white-box, we consider all possible box predictions whereas for black-box only the winning boxes after NMS are considered. This is explained in more detail in the following section.

\section{Experimental Evaluation}
\label{section:experiments}
In order to analyze the confidence calibration under different conditions, we use the COCO 2017 validation dataset \cite{Lin2014} with a random split of 70\% and 30\% for training and testing the calibration, respectively.

\subsection{Evaluation Protocol}
We perform both black-box and white-box calibration by following the evaluation protocol of \citep{Kueppers2020} and use their provided calibration framework. The final calibration results are obtained as an average over 20 independent training and testing results.
For inference, we use a pretrained RetinaNet \cite{Lin2017} and a Faster R-CNN \cite{Ren2015} model provided by the \textit{Detectron2} framework \cite{wu2019detectron2}. While the classification branch of the former model is trained by cross entropy loss, the latter one uses a focal loss that enables to focus on hard examples during training with low confidence. On the other hand, good predictions with high confidence are less weighted during training that in turn leads to less confident predictions \cite{Lin2017, Mukhoti2020}. Our experiments are restricted to the predictions of class \textit{person}.

To study the effect of non-maximum suppression, we apply different IoU thresholds to merge boxes denoted by $\text{NMS}@\{0.5, 0.75, 0.9\}$. In the white-box case without NMS, we use the raw predictions for measuring and performing calibration on the one hand. On the other hand, we further adopt $\text{\textit{top-}}k$ box selection where only $k$ bounding boxes with the highest confidence are kept using $k=1000$. This is the common case during inference to reduce low confidential and mostly redundant predictions. Following \cite{Kueppers2020}, the predictions of all models are obtained by inference with a probability threshold of $0.3$ which means discarding all predictions with a confidence score less than this threshold. As the relative amount of predictions per image with low confidence score is significantly higher than the relative amount of the remaining predictions, this probability threshold ensures that the D-ECE is not dominated by these low confidence samples.

For confidence calibration, we use multivariate histogram binning \cite{Zadrozny2001, Kueppers2020} for calibration as a fast and reliable calibration method. We also evaluate several setups with different subsets of box information to evaluate the effect of the used feature set. We either use the confidence only, also including the box centers $(\hat{c_x}, \hat{c_y})$ or box scales $(h, w)$, or we use all features for measuring and performing calibration. For the histogram-based calibration, we use 15 bins for confidence only, $N_k = 5$ bins for $(\hat{p}, \hat{c_x}, \hat{c_y})$ and $(\hat{p}, \hat{h}, \hat{w})$, and $N_k = 3$ when using all available features. In contrast, for D-ECE computation we use 20 bins for confidence only, $N_k = 8$ bins for $(\hat{p}, \hat{c_x}, \hat{c_y})$ and $(\hat{p}, \hat{h}, \hat{w})$, and $N_k = 5$ when using all available information. We increase the robustness of the D-ECE calculation by also neglecting bins with less than 8 samples \cite{Kueppers2020}.

\subsection{Results}
In Tables \ref{tab:ece:retinanet:precision} and \ref{tab:ece:faster_rcnn:precision}, the results for black-box and white-box calibration for RetinaNet and Faster R-CNN are presented, respectively. Three different IoU threshold values of $\tau = \{0.5, 0.6, 0.75\}$ are considered to match predictions with ground-truth annotations. In the tables, each cell presents the D-ECE for the baseline (without calibration) and the corresponding D-ECE after histogram-based calibration (HB). The Tables \ref{tab:ece:retinanet:precision} and \ref{tab:ece:faster_rcnn:precision} show the results of the black-box models with varying strength of NMS as well as the calibration results for the white-box case without NMS. The D-ECE is evaluated with different additional box information: The first column shows the confidence only calibration, the second and third columns the calibration with box centers and box scales, and the last columns show the results for the calibration with all box information considered. The best D-ECE scores are highlighted for each set of features and IoU value across all variants.

\begin{figure*}[htbp]
\centering
	\begin{subfigure}[b]{0.23\textwidth}
		\centering
		
		\begin{overpic}[width=\linewidth]{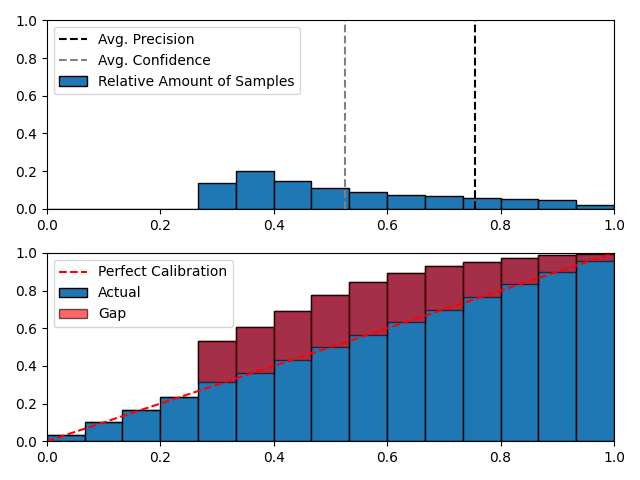}
		\put(85, 82){RetinaNet}
		\put(-6, 41){\rotatebox{90}{\tiny \% of samples}}
		\put(-6, 11){\rotatebox{90}{\tiny Precision}}
		\end{overpic}

		\subcaption{Uncalibrated white-box model with $\text{D-ECE}=22.913\%$}
		\label{fig:retinanet:conf_unc_whitebox}
	\end{subfigure}
	\begin{subfigure}[b]{0.23\textwidth}
		\centering
		
		\begin{overpic}[width=\linewidth]{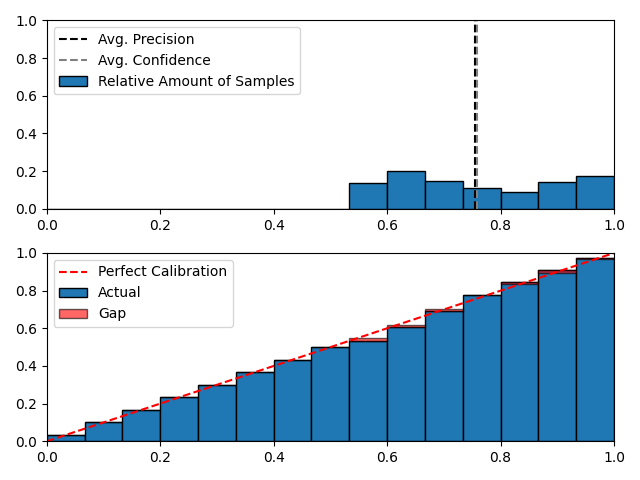}
		\end{overpic}
		
		\subcaption{Calibrated white-box model with $\text{D-ECE}=0.981\%$}
		\label{fig:retinanet:conf_cal_whitebox}
	\end{subfigure}
	\begin{subfigure}[b]{0.23\textwidth}
		\centering
		
		\begin{overpic}[width=\linewidth]{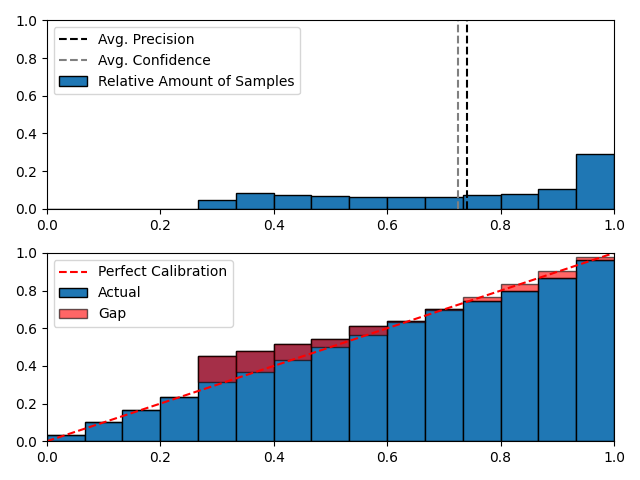}
		\put(75, 82){Faster R-CNN}
		\end{overpic}

		\subcaption{Uncalibrated white-box model with $\text{D-ECE}=4.198\%$}
		\label{fig:faster_rcnn:conf_unc_whitebox}
	\end{subfigure}
	\begin{subfigure}[b]{0.23\textwidth}
		\centering
		
		\begin{overpic}[width=\linewidth]{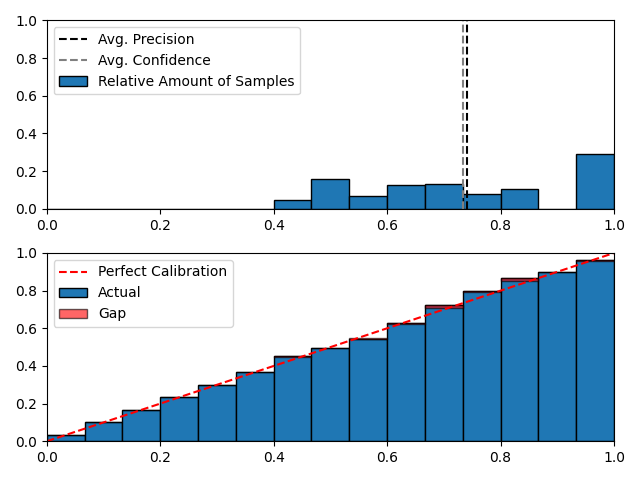}
		\end{overpic}
		
		\subcaption{Calibrated white-box model: $D-ECE=0.861\%$}
		\label{fig:faster_rcnn:conf_cal_whitebox}
	\end{subfigure}
	
	
	\begin{subfigure}[b]{0.23\textwidth}
		\centering
		
		\begin{overpic}[width=\linewidth]{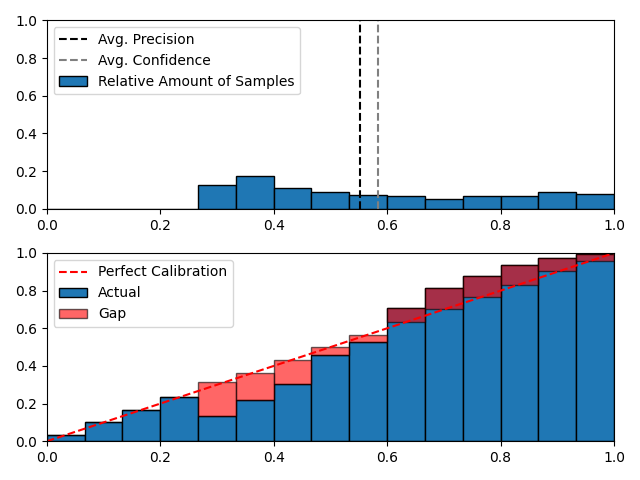}
		\put(-6, 41){\rotatebox{90}{\tiny \% of samples}}
		\put(-6, 11){\rotatebox{90}{\tiny Precision}}
		\end{overpic}
		
		\subcaption{Uncalibrated black-box model with $\text{D-ECE}=10.350\%$}
		\label{fig:retinanet:conf_unc_blackbox}
	\end{subfigure}
	\begin{subfigure}[b]{0.23\textwidth}
		\centering
		
		\begin{overpic}[width=\linewidth]{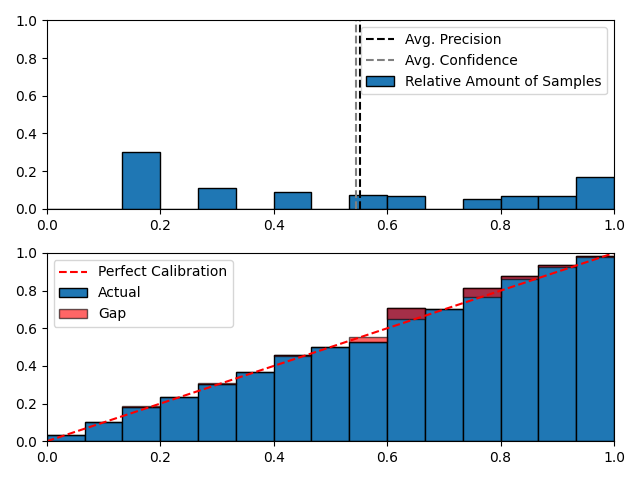}
		\end{overpic}
		
		\subcaption{Calibrated black-box model with $\text{D-ECE}=1.210\%$}
		\label{fig:retinanet:conf_cal_blackbox}
	\end{subfigure}
	\begin{subfigure}[b]{0.23\textwidth}
		\centering
		
		\begin{overpic}[width=\linewidth]{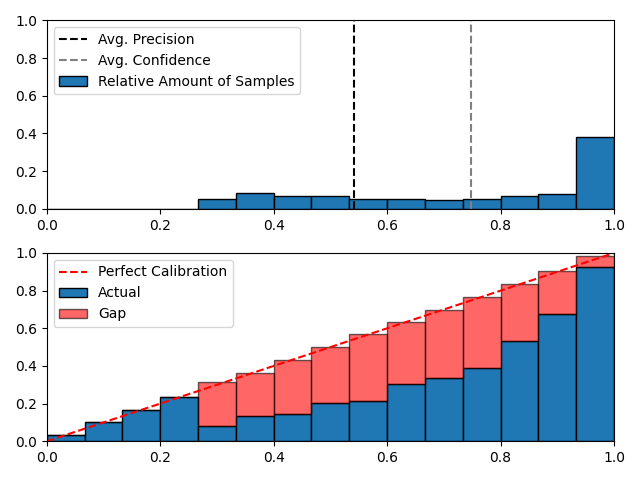}
		\end{overpic}
		
		\subcaption{Uncalibrated black-box model with $\text{D-ECE}=20.527\%$}
		\label{fig:faster_rcnn:conf_unc_blackbox}
	\end{subfigure}
	\begin{subfigure}[b]{0.23\textwidth}
		\centering
		
		\begin{overpic}[width=\linewidth]{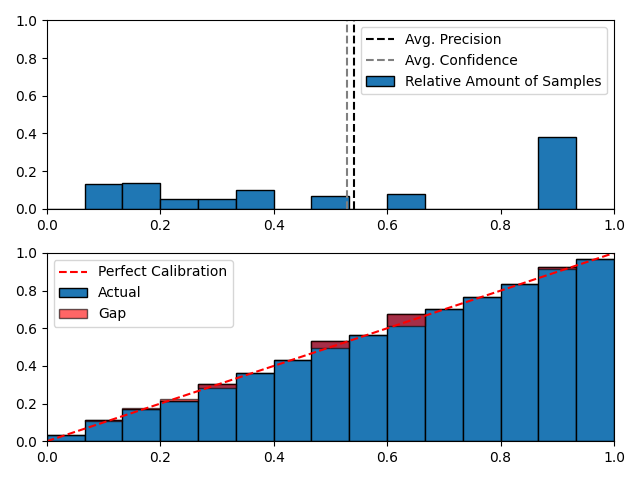}
		\end{overpic}
		
		\subcaption{Calibrated black-box model with $\text{D-ECE}=1.615\%$}
		\label{fig:faster_rcnn:conf_cal_blackbox}
	\end{subfigure}
	\caption{Confidence histograms and reliability diagrams of the miscalibration for RetinaNet (left) and Faster R-CNN (right) black-box (NMS@0.5) and white-box (without NMS)  models with IoU@0.6 before and after histogram-based calibration.}
	\label{fig:both:conf}
\end{figure*}

For Faster R-CNN, we observe that the white-box model calibrates consistently better by default than the black-box models in most cases. In contrast, we observe the opposite behavior for the RetinaNet model. Therefore, we further study the calibration properties of those networks by inspecting their reliability diagrams shown in Fig. \ref{fig:both:conf} for the black-box and white-box cases. The RetinaNet white-box model without NMS offers underconfident predictions which is a known property of models trained by focal loss \cite{Lin2017}. After NMS, a particular behavior can be observed in Fig. \ref{fig:retinanet:conf_unc_blackbox} with overconfident predictions in the low confidence interval ($\hat{p} < 0.5$) and underconfident predictions in the high confidence interval ($\hat{p} > 0.5$). Also, when comparing the calibrated results shown in Fig. \ref{fig:retinanet:conf_cal_whitebox} and \ref{fig:retinanet:conf_cal_blackbox}, it is evident that the calibration for the white-box model leads to a better D-ECE score. In contrast, Faster R-CNN outputs reasonably well calibrated predictions before NMS but is highly overconfident after NMS. Again, we observe that the white-box D-ECE score is much better compared to the black-box model \textbf{after} calibration has been applied. 

\begin{figure*}[htbp]
\centering
	\begin{subfigure}[b]{0.23\textwidth}
		\centering
		
		\begin{overpic}[width=\linewidth]{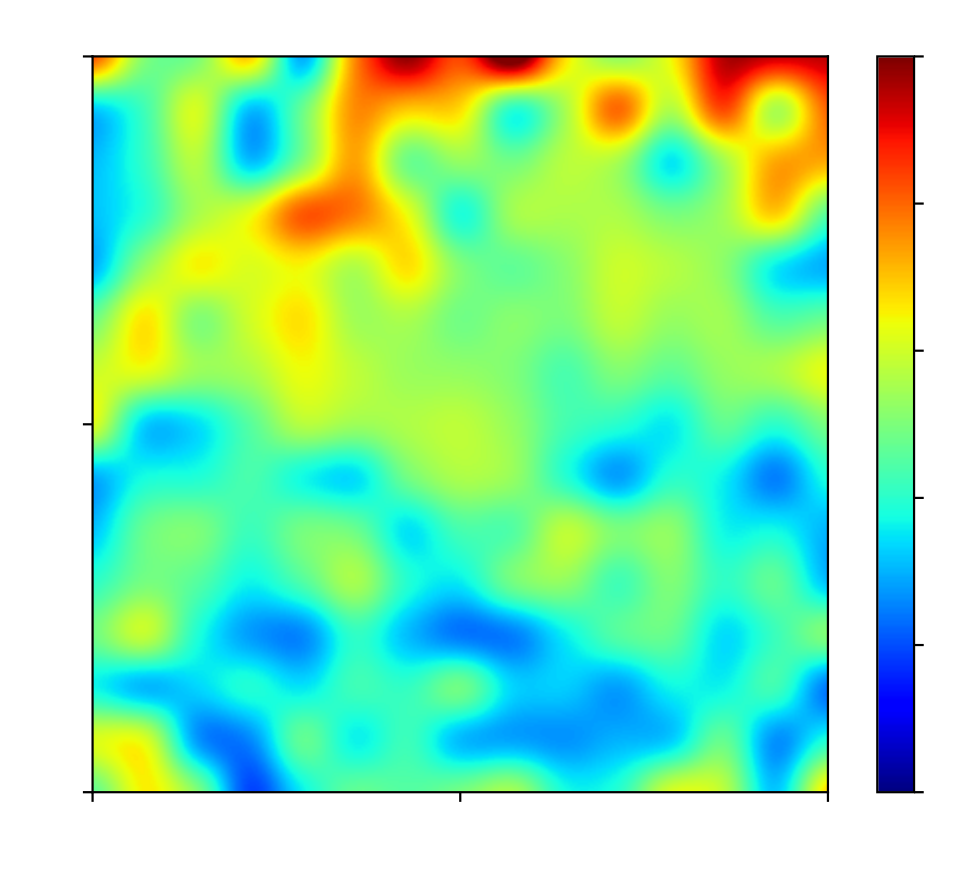}
		\put(85, 95){RetinaNet}
			\put(30, 86){\scriptsize{Default D-ECE}}
			\put(40, -2){\tiny{relative $c_x$}}
			\put(0, 35){\rotatebox{90}{\tiny{relative $c_y$}}}
			
			\put(2, 3){\tiny{0.0}}
			\put(45, 3){\tiny{0.5}}
			\put(83, 3){\tiny{1.0}}
			\put(1, 83){\tiny{1.0}}
			
			\put(97, 7){\tiny{0}}
			\put(97, 22){\tiny{1}}
			\put(97, 37){\tiny{2}}
			\put(97, 52){\tiny{3}}
			\put(97, 67.3){\tiny{4}}
			\put(97, 82.5){\tiny{5}}
			
			\put(90, 88){\tiny{1e-1}}
		\end{overpic}
		\subcaption{Uncalibrated white-box model with $\text{D-ECE}=22.992\%$}
	\end{subfigure}
	\begin{subfigure}[b]{0.23\textwidth}
		\centering
		\begin{overpic}[width=\linewidth]{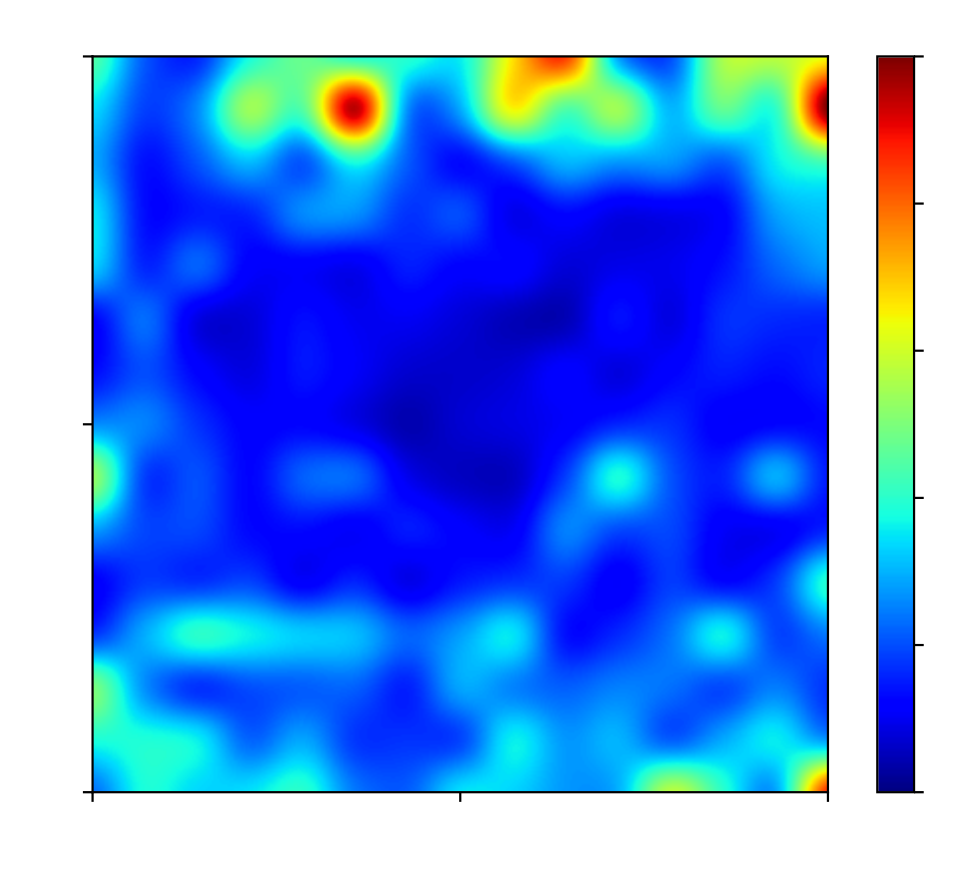}
			\put(18, 86){\scriptsize{After Histogram Binning}}
			\put(40, -2){\tiny{relative $c_x$}}
			\put(0, 35){\rotatebox{90}{\tiny{relative $c_y$}}}
			
			\put(2, 3){\tiny{0.0}}
			\put(45, 3){\tiny{0.5}}
			\put(83, 3){\tiny{1.0}}
			\put(1, 83){\tiny{1.0}}
			
			\put(97, 7){\tiny{0}}
			\put(97, 22){\tiny{1}}
			\put(97, 37){\tiny{2}}
			\put(97, 52){\tiny{3}}
			\put(97, 67.3){\tiny{4}}
			\put(97, 82.5){\tiny{5}}
			
			\put(90, 88){\tiny{1e-1}}
		\end{overpic}
		
		\subcaption{Calibrated white-box model with $\text{D-ECE}=5.671\%$}
	\end{subfigure}
	\begin{subfigure}[b]{0.23\textwidth}
		\centering
		
		\begin{overpic}[width=\linewidth]{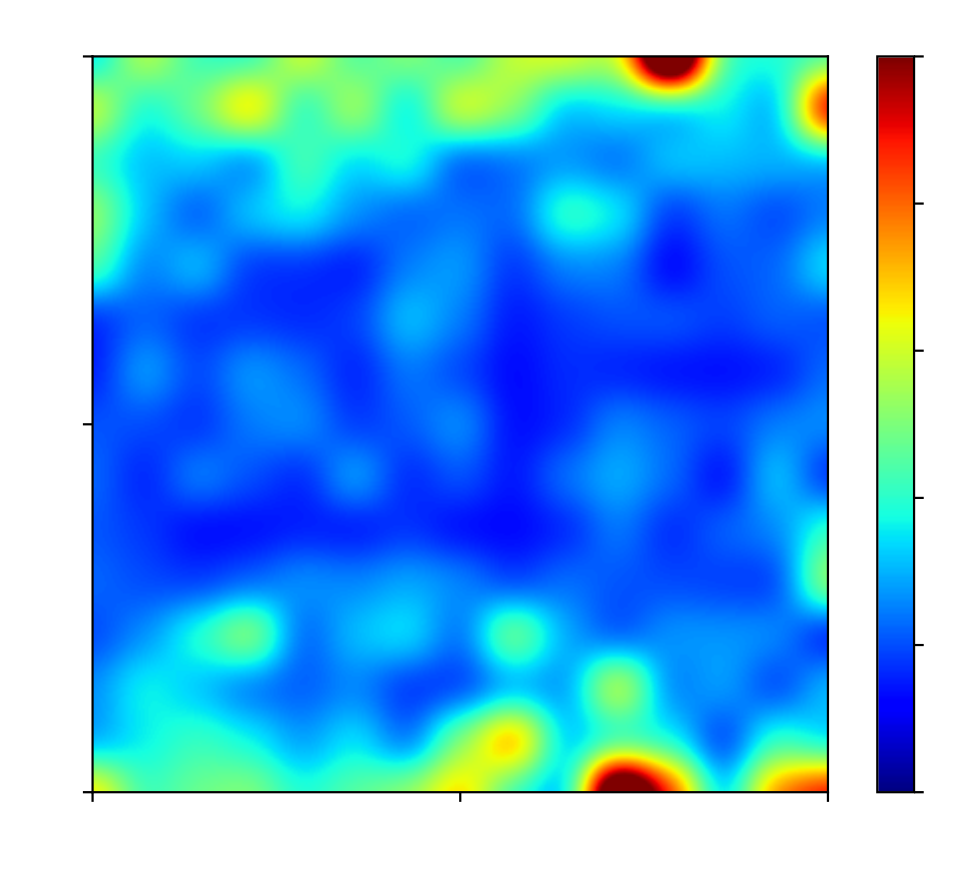}
		\put(75, 95){Faster R-CNN}
			\put(30, 86){\scriptsize{Default D-ECE}}
			\put(40, -2){\tiny{relative $c_x$}}
			\put(0, 35){\rotatebox{90}{\tiny{relative $c_y$}}}
			
			\put(2, 3){\tiny{0.0}}
			\put(45, 3){\tiny{0.5}}
			\put(83, 3){\tiny{1.0}}
			\put(1, 83){\tiny{1.0}}
			
			\put(97, 7){\tiny{0}}
			\put(97, 22){\tiny{1}}
			\put(97, 37){\tiny{2}}
			\put(97, 52){\tiny{3}}
			\put(97, 67.3){\tiny{4}}
			\put(97, 82.5){\tiny{5}}
			
			\put(90, 88){\tiny{1e-1}}
		\end{overpic}
		\subcaption{Uncalibrated white-box model with $\text{D-ECE}=7.631\%$}
	\end{subfigure}
	\begin{subfigure}[b]{0.23\textwidth}
		\centering
		\begin{overpic}[width=\linewidth]{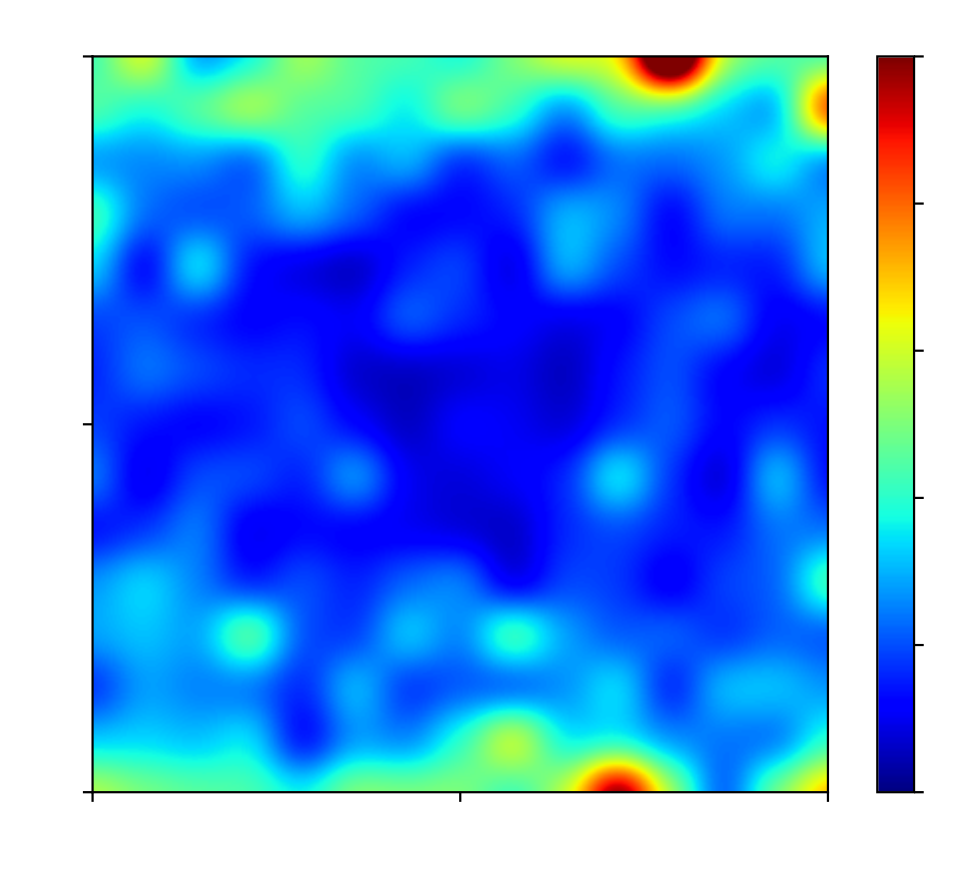}
			\put(18, 86){\scriptsize{After Histogram Binning}}
			\put(40, -2){\tiny{relative $c_x$}}
			\put(0, 35){\rotatebox{90}{\tiny{relative $c_y$}}}
			
			\put(2, 3){\tiny{0.0}}
			\put(45, 3){\tiny{0.5}}
			\put(83, 3){\tiny{1.0}}
			\put(1, 83){\tiny{1.0}}
			
			\put(97, 7){\tiny{0}}
			\put(97, 22){\tiny{1}}
			\put(97, 37){\tiny{2}}
			\put(97, 52){\tiny{3}}
			\put(97, 67.3){\tiny{4}}
			\put(97, 82.5){\tiny{5}}
			
			\put(90, 88){\tiny{1e-1}}
		\end{overpic}
		
		\subcaption{Calibrated white-box model with $\text{D-ECE}=5.998\%$}
	\end{subfigure}
	

	\begin{subfigure}[b]{0.23\textwidth}
		\centering	
		\begin{overpic}[width=\linewidth]{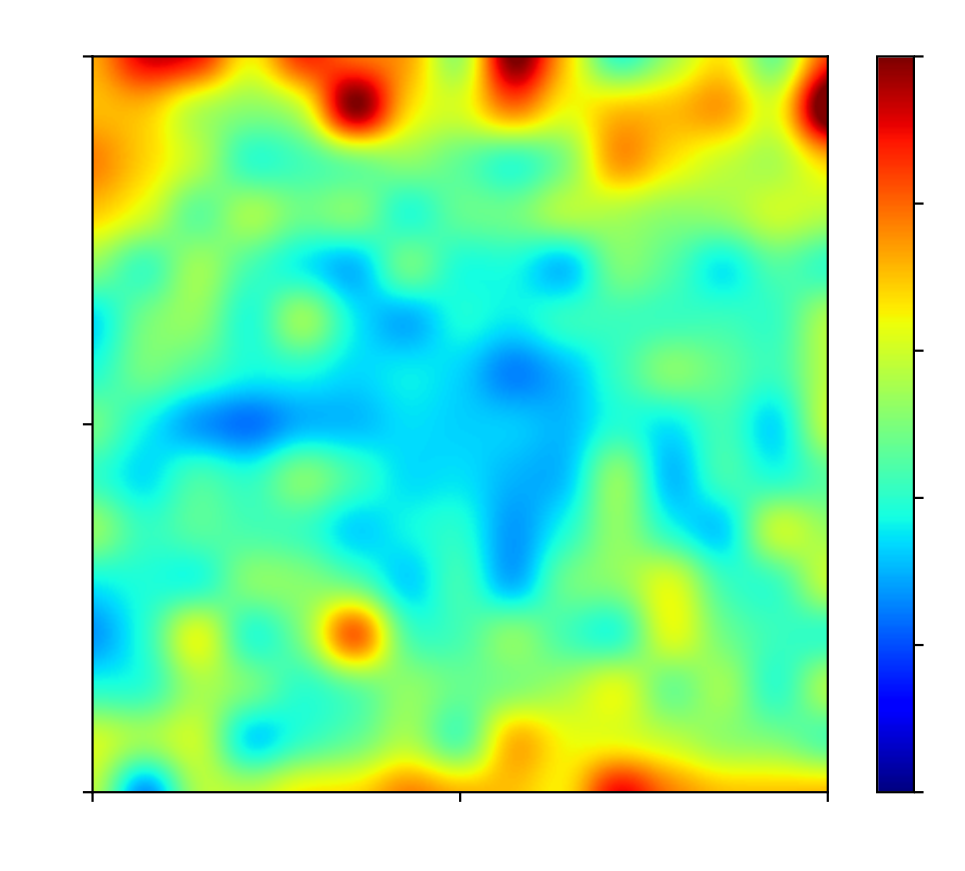}
			\put(30, 86){\scriptsize{Default D-ECE}}
			\put(40, -2){\tiny{relative $c_x$}}
			\put(0, 35){\rotatebox{90}{\tiny{relative $c_y$}}}
			
			\put(2, 3){\tiny{0.0}}
			\put(45, 3){\tiny{0.5}}
			\put(83, 3){\tiny{1.0}}
			\put(1, 83){\tiny{1.0}}
			
			\put(97, 7){\tiny{0}}
			\put(97, 22){\tiny{1}}
			\put(97, 37){\tiny{2}}
			\put(97, 52){\tiny{3}}
			\put(97, 67.3){\tiny{4}}
			\put(97, 82.5){\tiny{5}}
			
			\put(90, 88){\tiny{1e-1}}
		\end{overpic}
		\subcaption{Uncalibrated black-box model with $\text{D-ECE}=10.894\%$}
	\end{subfigure}
	\begin{subfigure}[b]{0.23\textwidth}
		\centering
		\begin{overpic}[width=\linewidth]{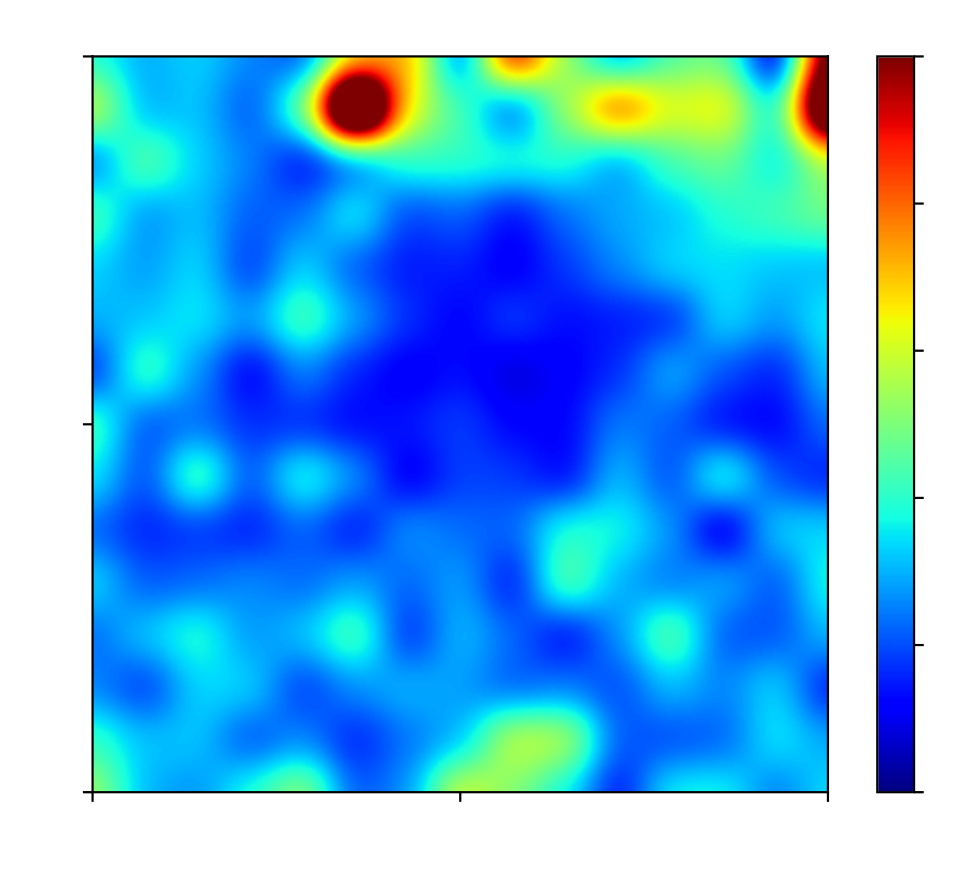}
			\put(18, 86){\scriptsize{After Histogram Binning}}
			\put(40, -2){\tiny{relative $c_x$}}
			\put(0, 35){\rotatebox{90}{\tiny{relative $c_y$}}}
			
			\put(2, 3){\tiny{0.0}}
			\put(45, 3){\tiny{0.5}}
			\put(83, 3){\tiny{1.0}}
			\put(1, 83){\tiny{1.0}}
			
			\put(97, 7){\tiny{0}}
			\put(97, 22){\tiny{1}}
			\put(97, 37){\tiny{2}}
			\put(97, 52){\tiny{3}}
			\put(97, 67.3){\tiny{4}}
			\put(97, 82.5){\tiny{5}}
			
			\put(90, 88){\tiny{1e-1}}
		\end{overpic}
		\subcaption{Calibrated black-box model with $\text{D-ECE}=6.620\%$}
	\end{subfigure}
	\begin{subfigure}[b]{0.23\textwidth}
		\centering	
		\begin{overpic}[width=\linewidth]{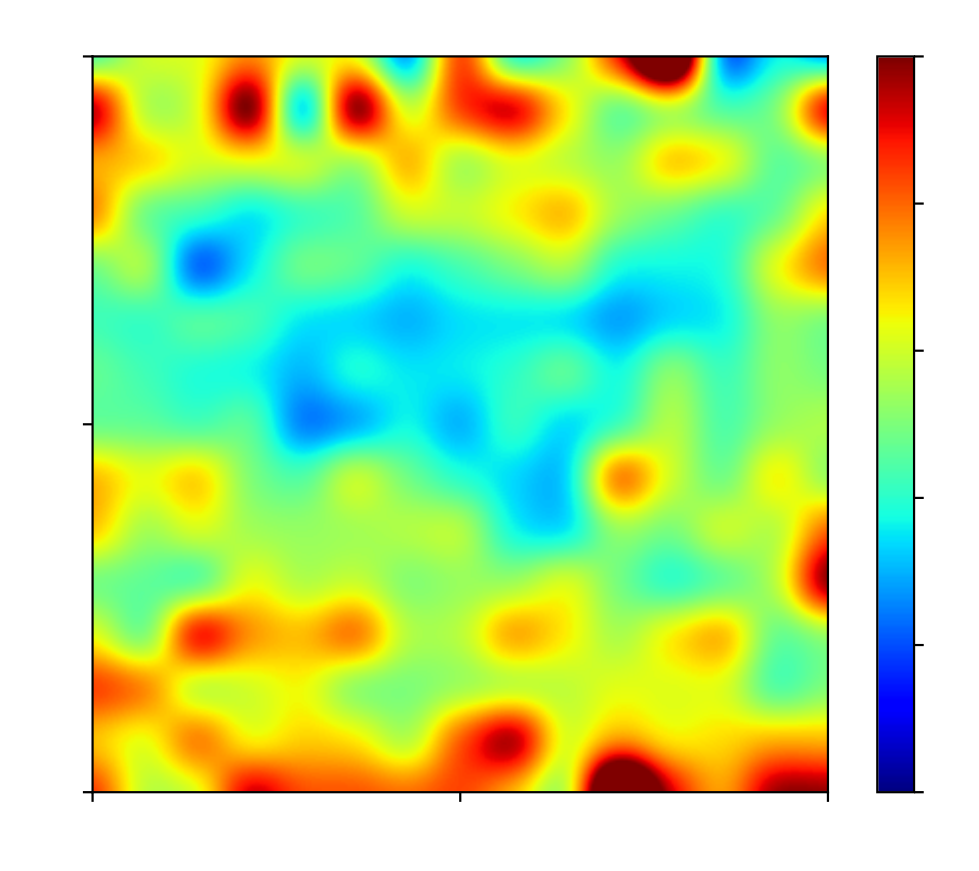}
			\put(30, 86){\scriptsize{Default D-ECE}}
			\put(40, -2){\tiny{relative $c_x$}}
			\put(0, 35){\rotatebox{90}{\tiny{relative $c_y$}}}
			
			\put(2, 3){\tiny{0.0}}
			\put(45, 3){\tiny{0.5}}
			\put(83, 3){\tiny{1.0}}
			\put(1, 83){\tiny{1.0}}
			
			\put(97, 7){\tiny{0}}
			\put(97, 22){\tiny{1}}
			\put(97, 37){\tiny{2}}
			\put(97, 52){\tiny{3}}
			\put(97, 67.3){\tiny{4}}
			\put(97, 82.5){\tiny{5}}
			
			\put(90, 88){\tiny{1e-1}}
		\end{overpic}
		\subcaption{Uncalibrated black-box model with $\text{D-ECE}=15.975\%$}
	\end{subfigure}
	\begin{subfigure}[b]{0.23\textwidth}
		\centering
		\begin{overpic}[width=\linewidth]{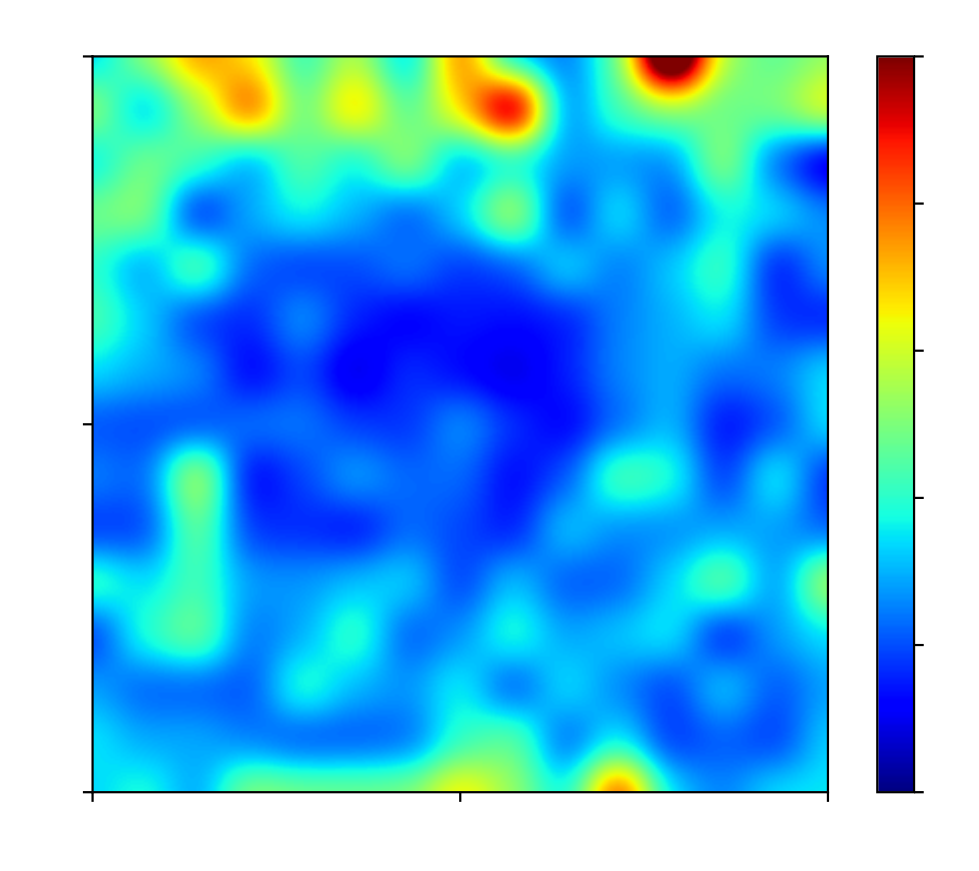}
			\put(18, 86){\scriptsize{After Histogram Binning}}
			\put(40, -2){\tiny{relative $c_x$}}
			\put(0, 35){\rotatebox{90}{\tiny{relative $c_y$}}}
			
			\put(2, 3){\tiny{0.0}}
			\put(45, 3){\tiny{0.5}}
			\put(83, 3){\tiny{1.0}}
			\put(1, 83){\tiny{1.0}}
			
			\put(97, 7){\tiny{0}}
			\put(97, 22){\tiny{1}}
			\put(97, 37){\tiny{2}}
			\put(97, 52){\tiny{3}}
			\put(97, 67.3){\tiny{4}}
			\put(97, 82.5){\tiny{5}}
			
			\put(90, 88){\tiny{1e-1}}
		\end{overpic}
		\subcaption{Calibrated black-box model with $\text{D-ECE}=7.156\%$}
	\end{subfigure}
	\caption{Position-dependent miscalibration of the RetinaNet (left) and Faster R-CNN (right) black-box (NMS@0.5) and white-box (without NMS) models with IoU@0.6, before and after the histogram-based calibration.}
	\label{fig:both:cx_cy}
\end{figure*}

\begin{table}[t]
	\centering
	\begin{tabular}{|c|c|c|c|c|}
		\hline
		& $(\hat{p})$ & $(\hat{p}, c_x, c_y)$ & $(\hat{p}, h, w)$ & full \\
		\hline
		NMS@0.5			& 0 & 1 & 29 & 528 \\
		\hline
		NMS@0.75		& 0 & 0 & 20 & 485 \\
		\hline
		NMS@0.9			& 0 & 0 & 12 & 435 \\
		\hline
		Without NMS		& 0 & 0 & 9 & 414 \\
		\hline
		Baseline & 20 & 256 & 256 & 1024 \\
		\hline
	\end{tabular}
	\caption{\label{tab:ece:faster_rcnn:bins} Amount of neglected bins within D-ECE calculation of the three black-box models and the white-box model for Faster R-CNN \cite{Ren2015, wu2019detectron2}. A similar amount of bins is also neglected during the examinations for RetinaNet \cite{Lin2017, wu2019detectron2}.}
\end{table}
We also study the effect of position-dependent miscalibration as in \cite{Kueppers2020}, shown in Fig. \ref{fig:both:cx_cy}. We compare the white-box and black-box models before and after calibration for each object detector. These figures allow to analyze if calibration is influenced by the position of predicted bounding boxes. All images show a tendency of higher miscalibration close to the borders. That may be caused by the difficulty of detecting objects correctly which are cropped out of the frame. However, this is of minor relevance considering that most of the positional discrepancies are mitigated after calibration in all cases.

As shown in Tables \ref{tab:ece:retinanet:precision} and \ref{tab:ece:faster_rcnn:precision}, the calibration for the white-box model performs better than the calibration for the black-box model for the first and second columns. The opposite happens when including the box scales into the computation of the D-ECE. Here, the black-box model with $\text{NMS@}0.5$ provides the best results. A possible explanation for this observation could be, that by increasing the NMS value, the number of samples also increases from 4,229 and 4,496 to 117,292 and 37,355 for RetinaNet and Faster R-CNN, respectively. As expected, the more we go in the white-box direction, the less predictions are discarded. Having more samples for the miscalibration computation also means that there are possibly more samples within each bin leading to a more robust miscalibration estimation \cite{Kumar2019}. As previously mentioned, bins with less than 8 samples are neglected for the computation of the D-ECE. The total amount of neglected bins for each configuration is illustrated in Table \ref{tab:ece:faster_rcnn:bins}.
Especially using all available information for calibration (full case), more and more bins are left out when going from white-box (bottom) to black-box (top) resulting in less bins contributing to the miscalibration score.

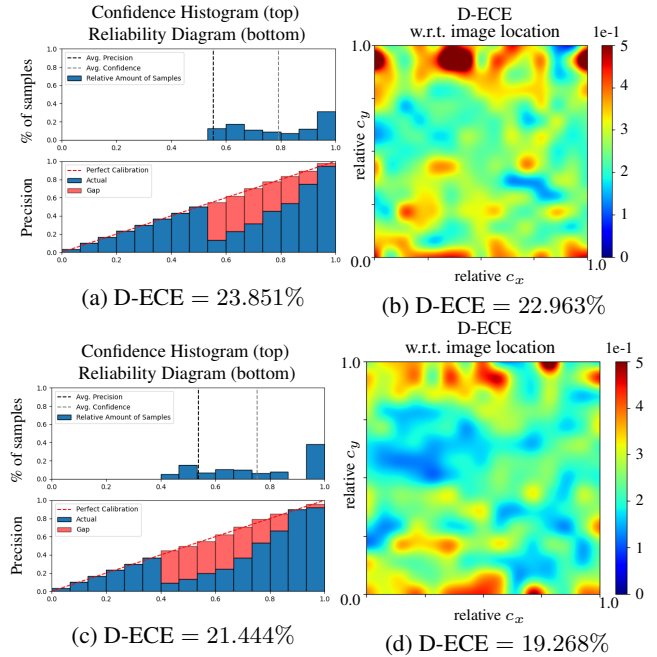
\begin{figure}[t]
	\centering
	\begin{subfigure}[b]{0.23\textwidth}
		\centering
		
		\begin{overpic}[width=\linewidth]{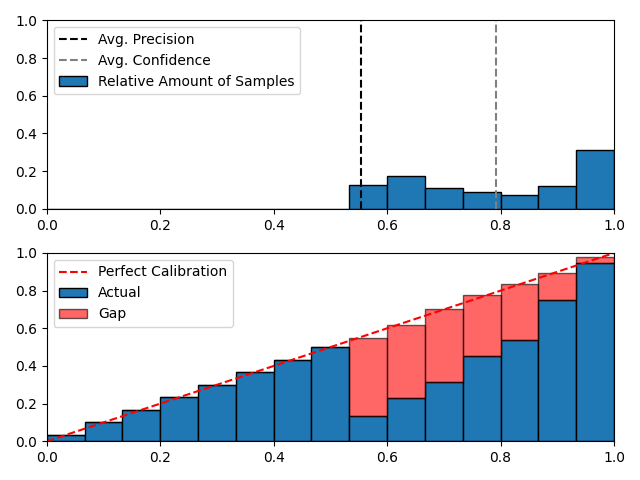}
			\put(17, 82){\scriptsize{Confidence Histogram (top)}}
			\put(16, 75){\scriptsize Reliability Diagram (bottom)}
			\put(-6, 41){\rotatebox{90}{\tiny \% of samples}}
			\put(-6, 11){\rotatebox{90}{\tiny Precision}}
		\end{overpic}
		
		\subcaption{$\text{D-ECE}  = 23.851\%$}
		\label{fig:retinanet:application:conf}
	\end{subfigure}
	\begin{subfigure}[b]{0.21\textwidth}
		\centering	
		\begin{overpic}[width=\linewidth]{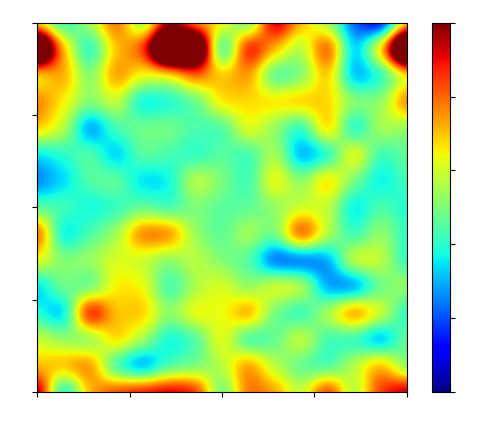}
			\put(37, 92){\scriptsize{D-ECE}}
			\put(20, 86){\scriptsize{w.r.t. image location}}
			\put(36, -1){\tiny{relative $c_x$}}
			\put(0, 35){\rotatebox{90}{\tiny{relative $c_y$}}}
			
			\put(-1, 4){\tiny{0.0}}
			\put(83, 3){\tiny{1.0}}
			\put(-1, 81){\tiny{1.0}}
			
			\put(95, 6){\tiny{0}}
			\put(95, 21){\tiny{1}}
			\put(95, 36.5){\tiny{2}}
			\put(95, 51.5){\tiny{3}}
			\put(95, 67){\tiny{4}}
			\put(95, 81){\tiny{5}}
			
			\put(85, 86){\tiny{1e-1}}
		\end{overpic}
		\subcaption{$\text{D-ECE} = 22.963\%$}
	\end{subfigure}
	\vspace{1em}

	\begin{subfigure}[b]{0.23\textwidth}
		\centering
		
		\begin{overpic}[width=\linewidth]{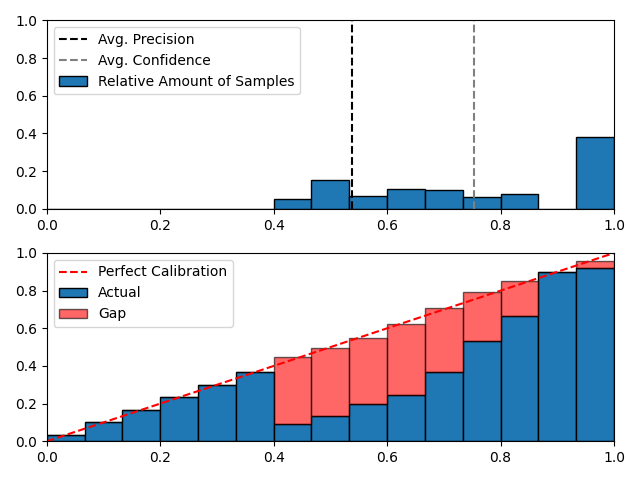}
			\put(17, 82){\scriptsize{Confidence Histogram (top)}}
			\put(16, 75){\scriptsize Reliability Diagram (bottom)}
			\put(-6, 41){\rotatebox{90}{\tiny \% of samples}}
			\put(-6, 11){\rotatebox{90}{\tiny Precision}}
		\end{overpic}
		
		\subcaption{$\text{D-ECE} = 21.444\%$}
		\label{fig:faster_rcnn:application:conf}
	\end{subfigure}
	\begin{subfigure}[b]{0.23\textwidth}
		\centering
		\begin{overpic}[width=\linewidth]{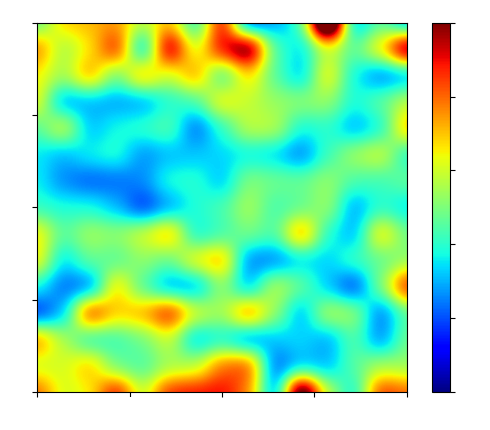}
			\put(37, 92){\scriptsize{D-ECE}}
			\put(20, 86){\scriptsize{w.r.t. image location}}
			\put(36, -1){\tiny{relative $c_x$}}
			\put(0, 35){\rotatebox{90}{\tiny{relative $c_y$}}}
			
			\put(-1, 4){\tiny{0.0}}
			\put(83, 3){\tiny{1.0}}
			\put(-1, 81){\tiny{1.0}}
			
			\put(95, 6){\tiny{0}}
			\put(95, 21){\tiny{1}}
			\put(95, 36.5){\tiny{2}}
			\put(95, 51.5){\tiny{3}}
			\put(95, 67){\tiny{4}}
			\put(95, 81){\tiny{5}}
			
			\put(85, 86){\tiny{1e-1}}
		\end{overpic}
		\subcaption{$\text{D-ECE} = 19.268 \%$}
	\end{subfigure}	
	\caption{Confidence histogram and reliability diagram (left) and position-dependent heatmap (right) for RetinaNet \cite{Lin2017, wu2019detectron2} (top row) and Faster RCNN \cite{Ren2015, wu2019detectron2} (bottom row) after white-box calibration and then further application of non-maximum suppression with NMS@0.5.}
	\label{fig:retinanet:application}
\end{figure}

A critical question arises how to integrate white-box calibration into the object detection pipeline. As demonstrated in the previous results, NMS has a significant impact in the calibration affecting the precision as well as the confidence scores of the detections. It has been shown that NMS has the potential to degrade the calibration results. Therefore, we investigate the calibration properties of the detection models that are processed by a NMS with histogram-based calibration beforehand. The results are shown in Fig. \ref{fig:retinanet:application}: It can be seen that calibration before NMS leads to higher miscalibration as the confidence is calibrated before NMS as well. However, as NMS also affects the precision, the detection model gets too overconfident in both cases.
In order to preserve good calibrations from the white-box method, alternative box suppression methods should be investigated. One option would be to integrate the confidence calibration with the box merging strategies compared by \cite{roza2020assessing}, such as \textit{weighted box fusion} and \textit{variance voting} and test how such methods influence the model calibration.

\begin{table*}
\centering
\begin{subtable}{0.45\textwidth}
	\begin{tabular}{|c|c|c|c|c|}
	\hline
				& $(\hat{p})$ & $(\hat{p}, c_x, c_y)$ & $(\hat{p}, h, w)$ & full \\
	\hline
	IoU@0.5		& & & &	\\
	Baseline		& 16.200 & 12.004 & 14.963 & 13.478 \\
	HB			& 1.636 & 6.335 & \textbf{2.775} & \textbf{5.391} \\
	\hline
	IoU@0.6		& & & &	\\
	Baseline		& 20.862 & 15.303 & 18.743 & 16.546 \\
	HB			& 1.673 & 6.091 & \textbf{2.991} & \textbf{5.743} \\
	\hline
	IoU@0.75		& & & &	\\
	Baseline		& 31.659 & 24.684 & 28.864 &	25.765 \\
	HB			& 1.436 & 6.095 & \textbf{3.110} & 5.704 \\
	\hline
	\end{tabular}
	\subcaption{\label{tab:ece:retinanet:precision:a} Black-box calibration with NMS@0.5, $|D|=4,229$.}
\end{subtable}
\begin{subtable}{0.45\textwidth}
	\begin{tabular}{|c|c|c|c|c|}
	\hline
				& $(\hat{p})$ & $(\hat{p}, c_x, c_y)$ & $(\hat{p}, h, w)$ & full  \\

	\hline
	IoU@0.5		& & & &	\\
	Baseline		& 15.448 & 15.750 & 14.246 & 12.586	\\
	HB			& 1.388 & 6.123 & 4.252 & 6.665	\\
	\hline
	IoU@0.6		& & & &	\\
	Baseline		&  3.435 & 7.486 & 6.710 & 7.064	\\
	HB			& 1.441 & 6.273 & 4.192 & 6.444 \\
	\hline
	IoU@0.75		& & & &	\\
	Baseline		& 20.980 & 20.840 & 20.041 &	17.504 \\
	HB			& 1.227 & \textbf{4.847} & 3.315 & \textbf{4.974} \\
	\hline
	\end{tabular}
	\subcaption{\label{tab:ece:retinanet:precision:b} Black-box calibration NMS@0.75, $|D|=7,923$}
\end{subtable}
\begin{subtable}{0.45\textwidth}
	\begin{tabular}{|c|c|c|c|c|}
	\hline
				& $(\hat{p})$ & $(\hat{p}, c_x, c_y)$ & $(\hat{p}, h, w)$ & full  \\

	\hline
	IoU@0.5		& & & &	\\
	Baseline		& 30.748 &	30.672 & 30.436 & 29.427\\
	HB			& 1.212 & 5.290 & 3.671 & 6.686 \\
	\hline
	IoU@0.6		& & & &	\\
	Baseline		& 21.773 & 21.954 & 21.612 &	 21.350 \\
	HB			& 1.195 & 5.717 & 3.981 & 7.647 \\
	\hline
	IoU@0.75		& & & &	\\
	Baseline		&  3.057 &	6.907 &  8.489 & 10.143 \\
	HB			& 1.367  & 5.675 & 4.468 & 7.847 \\
	\hline
	\end{tabular}
	\subcaption{\label{tab:ece:retinanet:precision:c} Black-box calibration with NMS@0.9, $|D|=20,005$}
\end{subtable}
\begin{subtable}{0.45\textwidth}
	\begin{tabular}{|c|c|c|c|c|}
	\hline
				& $(\hat{p})$ & $(\hat{p}, c_x, c_y)$ & $(\hat{p}, h, w)$ & full \\

	\hline
	IoU@0.5		& & & &	\\
	Baseline		& 28.027 & 28.127 & 28.014 & 28.176	\\
	HB			& \textbf{0.855} & \textbf{4.947} & 2.895 & 6.114 \\
	\hline
	IoU@0.6		& & & &	\\
	Baseline		& 23.097 & 23.290 & 23.118 &	23.482 \\
	HB			& \textbf{1.033}  & \textbf{5.331} & 3.306 & 6.912 \\
	\hline
	IoU@0.75		& & & &	\\
	Baseline		& 8.487 & 10.190 & 10.190 & 11.992 \\
	HB			& \textbf{1.132} & 5.892 & 4.207 & 8.266 \\
	\hline
	\end{tabular}
	\subcaption{\label{tab:ece:retinanet:precision:d} White-box calibration without NMS, $|D|=117,292$}
\end{subtable}
\caption{D-ECE results (\%) for RetinaNet \cite{Lin2017, wu2019detectron2} for different IoU scores. Each column shows the baseline D-ECE and the calibrated one using histogram-based (HB) calibration with different subsets using either confidence only $\hat{p}$, including the box centers $(c_x, c_y)$, box scales $(h, w)$ or using all features. Note that comparing the D-ECE scores of columns to each other is not applicable since different subsets of data have been used for D-ECE measurement and calibration.}
\label{tab:ece:retinanet:precision}
\end{table*}


\begin{table*}
\centering
\begin{subtable}{0.45\textwidth}
	\begin{tabular}{|c|c|c|c|c|}
	\hline
				& $(\hat{p})$ & $(\hat{p}, c_x, c_y)$ & $(\hat{p}, h, w)$ & full \\
	\hline
	IoU@0.5		& & & &	\\
	Baseline		& 7.781 & 9.060 & 7.829 & 7.168 \\
	HB			& 1.789 & 6.186 & \textbf{2.947} & \textbf{4.960} \\
	\hline
	IoU@0.6		& & & &	\\
	Baseline		& 9.370 & 10.041 & 9.033 & 7.810 \\
	HB			& 1.564 & 6.075 & \textbf{3.105} & \textbf{5.142} \\
	\hline
	IoU@0.75		& & & &	\\
	Baseline		& 31.659 & 24.684 & 28.864 &	25.765 \\
	HB			& 1.436 & 6.095 & 3.110 & 5.704 \\
	\hline
	\end{tabular}
	\subcaption{\label{tab:ece:faster_rcnn:precision:a} Black-box calibration with NMS@0.5, $|D|=4,496$.}
\end{subtable}
\begin{subtable}{0.45\textwidth}
	\begin{tabular}{|c|c|c|c|c|}
	\hline
				& $(\hat{p})$ & $(\hat{p}, c_x, c_y)$ & $(\hat{p}, h, w)$ & full  \\

	\hline
	IoU@0.5		& & & &	\\
	Baseline		& 7.597 & 9.927 & 10.828 & 9.804	\\
	HB			& 1.523 & 6.968 & 3.778 & 6.134 \\
	\hline
	IoU@0.6		& & & &	\\
	Baseline		&  16.100 & 15.226 & 16.417 & 14.933	\\
	HB			& 1.343 & 6.323 & 3.490 & 5.610 \\
	\hline
	IoU@0.75		& & & &	\\
	Baseline		& 34.634 & 32.535 & 31.861 &	27.883 \\
	HB			& 1.123 & \textbf{4.878} & \textbf{3.018} & \textbf{4.996} \\
	\hline
	\end{tabular}
	\subcaption{\label{tab:ece:faster_rcnn:precision:b} Black-box calibration with NMS@0.75, $|D|=7,231$.}
\end{subtable}
\begin{subtable}{0.45\textwidth}
	\begin{tabular}{|c|c|c|c|c|}
	\hline

				& $(\hat{p})$ & $(\hat{p}, c_x, c_y)$ & $(\hat{p}, h, w)$ & full \\
	\hline
	IoU@0.5		& & & &	\\
	Baseline		& 7.323 &	10.431 & 10.042 & 10.318 \\
	HB			& 1.354  & 6.697 & 4.062 & 7.121 \\
	\hline
	IoU@0.6		& & & &	\\
	Baseline		& 7.499 & 10.328 & 11.622 &	 11.630 \\
	HB			& 1.184 & 6.383 & 4.050 & 7.141 \\
	\hline
	IoU@0.75		& & & &	\\
	Baseline		& 25.689 &	25.539  & 25.792 & 25.002 \\
	HB			& 1.139    & 5.478 & 4.126 & 6.908 \\
	\hline
	\end{tabular}
	\subcaption{\label{tab:ece:faster_rcnn:precision:c} Black-box calibration with NMS@0.9, $|D|=17,742$.}
\end{subtable}
\begin{subtable}{0.45\textwidth}
	\begin{tabular}{|c|c|c|c|c|}
	\hline
				& $(\hat{p})$ & $(\hat{p}, c_x, c_y)$ & $(\hat{p}, h, w)$ & full \\
	\hline
	IoU@0.5		& & & &	\\
	Baseline		& 6.914 & 9.619 & 8.638 & 10.061	\\
	HB			& \textbf{1.038} & \textbf{5.234} & 3.206 & 6.239 \\
	\hline
	IoU@0.6		& & & &	\\
	Baseline		& 4.592 & 7.720 & 8.540 & 9.548 \\
	HB			& \textbf{1.099} & \textbf{5.523} & 3.603 & 6.959 \\
	\hline
	IoU@0.75		& & & &	\\
	Baseline		& 13.067 & 13.883 & 15.658 & 16.462 \\
	HB			& \textbf{0.999} & 5.996 & 4.505 & 8.652 \\
	\hline
	\end{tabular}
	\subcaption{\label{tab:ece:faster_rcnn:precision:d} White-box calibration without NMS, $|D|=37,355$. }
\end{subtable}
\caption{\label{tab:ece:faster_rcnn:precision} D-ECE results (\%) for Faster R-CNN \cite{Ren2015, wu2019detectron2} before and after histogram-based (HB) calibration using different IoU thresholds for NMS. The structure of this table is comparable to Tab. \ref{tab:ece:retinanet:precision}.}
\end{table*}

\section{Conclusion}
\label{section:conclusion}

In this paper, we analyzed the influence of box suppression methods on confidence calibration for object detection models. To do so, we adapt models without box suppression methods denoted as white-box models, contrasting to the black-box approach commonly suggested. We performed histogram-based calibration for both black-box and white-box scenarios on the COCO dataset. We found that the initial calibration of detection models is highly impacted by NMS. Additionally, we observed that calibration also depends on the architecture of the object detection model. For RetinaNet, the model predictions are underconfident before applying NMS whereas, for Faster R-CNN, the white-box model outputs quite well calibrated detections that become overconfident after NMS.

Knowing that the miscalibration not only depends on the classification outputs but also on the regression output for the bounding boxes, we performed histogram-based calibration using different subsets of the output data. For the confidence only and $(\hat{p}, c_x, c_y)$ case, the white-box model outperforms the black-box models while the black-box models present slightly better results on the other scenarios.

While the white-box calibration has given good results, the most effective integration of white-box calibration methods in existing object detectors utilizing NMS remains as an open issue. As shown by the results in this paper, the NMS layer affects the results by giving different calibration profiles before and after the suppression. Corroborating with further results presented in this paper, the calibrated detections obtained by the white-box models deteriorated after NMS for both RetinaNet and Faster R-CNN. However, we think this problem can be solved by using other suppression methods which consider a larger set of the overall better calibrated boxes than NMS.

For future work we suggest alternative applications to the standard NMS method to verify if they can lead to better calibrated object detectors. One option would be to integrate the confidence calibration with box merging strategies compared by \cite{roza2020assessing}, such as \textit{box averaging}, \textit{weighted box fusion} or \textit{variance voting}. 

\section*{Aknowledgements}

This work was funded by the Bavarian Ministry for Economic Affairs, Regional Development and Energy as part of a project to support the thematic development of the Institute for Cognitive Systems and within the Intel Collaborative Research Institute Safe Automated Vehicles.

\begin{quote}
\begin{small}
\bibliography{bibliography}
\end{small}
\end{quote}

\end{document}